\documentclass[conference]{IEEEtran}

\makeatletter

\def\ps@IEEEtitlepagestyle{%
\def\@oddfoot{\mycopyrightnotice}%
\def\@evenfoot{}%
}
\def\mycopyrightnotice{%
{\footnotesize 978-1-6654-7087-2/22/\$31.00~\copyright~2022 IEEE\hfill}
\gdef\mycopyrightnotice{}
}

\usepackage{blindtext}
\usepackage{eso-pic}
\IEEEoverridecommandlockouts
\usepackage{cite}
\usepackage{amsmath,amssymb,amsfonts}
\usepackage{algorithmic}
\usepackage{graphicx}
\usepackage{float}
\usepackage{hyperref}
\usepackage{placeins}
\usepackage{textcomp}
\usepackage{multirow}
\usepackage{xcolor}
\def\BibTeX{{\rm B\kern-.05em{\sc i\kern-.025em b}\kern-.08em
	T\kern-.1667em\lower.7ex\hbox{E}\kern-.125emX}}

\usepackage{eso-pic}
\newcommand\AtPageUpperMyright[1]{\AtPageUpperLeft{%
	\put(\LenToUnit{0.17\paperwidth},\LenToUnit{-2cm}){%
		\parbox{0.9\textwidth}{\raggedleft\fontsize{8}{11}\selectfont #1}}%
}}%
\newcommand{\conf}[1]{%
\AddToShipoutPictureBG*{%
	\AtPageUpperMyright{#1}
}
}

\newcommand*{\dittoclosing}{ \raisebox{-0.5ex}{''} }

\makeatletter
\newcommand{\linebreakand}{%
\end{@IEEEauthorhalign}
\hfill\mbox{}\par
\mbox{}\hfill\begin{@IEEEauthorhalign}
}
\makeatother

\begin{document}
\title{\vspace*{1cm} Automatic Tuberculosis and COVID-19 cough classification using deep learning \\
		%
		%
}

\author{\IEEEauthorblockN{Madhurananda Pahar}
	\IEEEauthorblockA{
	\textit{Department of Electrical and Electronic Engineering, } \\
			\textit{Stellenbosch University,}\\
			Stellenbosch, South Africa. \\
			mpahar@sun.ac.za}
	\and
	\IEEEauthorblockN{Marisa Klopper}
	\IEEEauthorblockA{
	    \textit{Division of Molecular Biology}\\
	        \textit{and Human Genetics,}\\
			\textit{Stellenbosch University,}\\
			Stellenbosch, South Africa. \\
			marisat@sun.ac.za}
	\and
	\IEEEauthorblockN{Byron Reeve}
	\IEEEauthorblockA{
	    \textit{Division of Molecular Biology}\\
	    \textit{and Human Genetics,}\\
			\textit{Stellenbosch University,}\\
			Stellenbosch, South Africa. \\
			byronreeve@sun.ac.za}
	\linebreakand 
	\IEEEauthorblockN{Rob Warren}
	\IEEEauthorblockA{
	    \textit{Division of Molecular Biology }\\
	    \textit{and Human Genetics,}\\
			\textit{Stellenbosch University,}\\
			Stellenbosch, South Africa. \\
			rw1@sun.ac.za}
	\and 
	\IEEEauthorblockN{Grant Theron}
	\IEEEauthorblockA{
	    \textit{Division of Molecular Biology and Human Genetics,}\\
			\textit{Stellenbosch University,}\\
			Stellenbosch, South Africa. \\
			gtheron@sun.ac.za}
	\and 
	\IEEEauthorblockN{Andreas Diacon}
	\IEEEauthorblockA{
	\textit{TASK Applied Science,} \\
			Cape Town, South Africa. \\
			ahd@task.org.za}
	\linebreakand 
	\IEEEauthorblockN{Thomas Niesler}
	\IEEEauthorblockA{
	\textit{Department of Electrical and Electronic Engineering,} \\
			\textit{Stellenbosch University,}\\
			Stellenbosch, South Africa. \\
			trn@sun.ac.za}
}

\maketitle
\conf{\textit{  Proc. of the International Conference on Electrical, Computer and Energy Technologies (ICECET 2022) \\ 
		20-22 July 2022, Prague-Czech Republic}}

\begin{abstract}
	We present a deep learning based automatic cough classifier which can discriminate tuberculosis (TB) coughs from COVID-19 coughs and healthy coughs. 
	Both TB and COVID-19 are respiratory diseases, contagious, have cough as a predominant symptom and claim thousands of lives each year. 
	The cough audio recordings were collected at both indoor and outdoor settings and also uploaded using smartphones from subjects around the globe, thus containing various levels of noise. 
	This cough data include 1.68 hours of TB coughs, 18.54 minutes of COVID-19 coughs and 1.69 hours of healthy coughs from 47 TB patients, 229 COVID-19 patients and 1498 healthy patients and were used to train and evaluate a CNN, LSTM and Resnet50. 
	These three deep architectures were also pre-trained on 2.14 hours of sneeze, 2.91 hours of speech and 2.79 hours of noise for improved performance. 
	The class-imbalance in our dataset was addressed by using SMOTE data balancing technique and using performance metrics such as F1-score and AUC. 
	Our study shows that the highest F1-scores of 0.9259 and 0.8631 have been achieved from a pre-trained Resnet50 for two-class (TB vs COVID-19) and three-class (TB vs COVID-19 vs healthy) cough classification tasks, respectively. 
	The application of deep transfer learning has improved the classifiers' performance and makes them more robust as they generalise better over the \mbox{cross-validation} folds. 
	Their performances exceed the TB triage test requirements set by the world health organisation (WHO). 
	The features producing the best performance contain higher order of MFCCs suggesting that the differences between TB and COVID-19 coughs are not perceivable by the human ear. 
	This type of cough audio classification is non-contact, cost-effective and can easily be deployed on a smartphone, thus it can be an excellent tool for both TB and COVID-19 screening. 
	
\end{abstract}


\vspace{5pt}

\begin{IEEEkeywords}
	tuberculosis, COVID-19, cough, transfer learning, deep learning, Resnet50
\end{IEEEkeywords}

\section{Introduction}
\label{sec:introduction}

Tuberculosis (TB) is a bacterial infectious disease which affects the human lungs, prevalent in low-income settings and 95\% of all TB cases are reported in developing countries 
\cite{TB-WHO, floyd2018global}. 
Modern diagnostic tests are costly as they rely on special equipment and laboratory procedure 
\cite{dewan2006feasibility,bwanga2009direct,konstantinos2010testing}. 
Suspected patients are tested when they show the symptom criteria of TB investigation and the results indicate that most of them cough due to other lung ailments; in fact most of those TB-suspected patients do not suffer from TB \cite{chang2008chronic}. 

COVID-19 (\textbf{CO}rona \textbf{VI}rus \textbf{D}isease of 20\textbf{19}) was declared as a global pandemic on February 11, 2020 by the World Health Organisation (WHO). 
At the time of writing, there are 513.9 million COVID-19 global cases and sadly, the pandemic has claimed the life of 6.2 million \cite{johnCOVID19}. 
Thus, many suspected TB patients are very likely to be suffering from COVID-19 in developing countries and experimental evidence suggests that healthy people cough less than those who are sick from lung ailments \cite{botha2018detection}. 
Therefore, there is a need for automated non-contact, low-cost, easily-accessible tools for both TB and COVID-19 screening based on cough audio. 

One of the major symptoms of respiratory diseases like TB and COVID-19 is a cough \cite{carfi2020persistent,wang2020clinical}. 
Depending on the nature of the respiratory disease, the airway is to be either blocked or restricted and
this can affect the acoustic properties of the coughs, 
thus enabling the cough audio to be used by machine learning algorithms in many studies including our own \cite{pahar2020covid, pahar2021tb, pahar2022covid} for discriminating both TB \cite{infante2017use} and COVID-19 \cite{laguarta2020covid} from healthy coughs. 
As TB is mostly found in developing countries, the efforts to collect TB coughs 
are rare, thus TB cough data are small and not publicly available. 
Successful studies \cite{tracey2011cough, botha2018detection, pahar2021tb} have experimentally found that shallow classifiers such as a multilayer perceptron (MLP) or logistic regression (LR) model works well in detecting TB in cough audio. 
However, COVID-19 cough data are widely available \cite{orlandic2020coughvid, sharma2020coswara, Schuller21-TI2} and many recent studies have successfully applied neural networks \cite{miranda2021machine} including deep neural network (DNN) classifiers to detect COVID-19 in cough audio \cite{imran2020ai4covid, tena2022automated, laguarta2020covid}.

In this study, we present a deep learning based automatic cough classifier which discriminates TB coughs from \mbox{COVID-19} coughs. 
We have used both public and private datasets and synthetic minority over-sampling technique (SMOTE) to create new datapoints to balance the datasets, as COVID-19 coughs are under-represented in our datasets. 
We have also used both AUC (area under the ROC curve) and F1-score as the performance metric for our three DNN classifiers: CNN, LSTM \& Resnet50 and evaluated them using nested cross-validation to make the best use of our datasets. 
The highest F1-score of 0.9042 has been achieved from a Resnet50 classifier in discriminating TB coughs from \mbox{COVID-19} coughs. 
Inspired by our previous research \cite{pahar2022covid}, 
we have made use of sneeze, speech and noise to pre-train these three deep architectures as well. 
This has improved the F1-score of this two-class classification task to 0.9259 with more robust performance across the cross-validation folds. 
The corresponding AUC has been 0.9245 with a 96\% sensitivity at 80\% specificity, exceeding the TB triage test requirement of 90\% sensitivity at 70\% specificity determined by WHO. 
We have further investigated these three DNN classifiers' performances in a three-class classification task, where we added healthy coughs as the third class. 
Initially, an \mbox{F1-score} of 0.8578  has been achieved from the Resnet50 and it has been improved to 0.8631 from the same architecture in discriminating TB, COVID-19 and healthy coughs after applying the transfer learning. 

Section \ref{sec:data} will detail the datasets used for pre-training the DNN classifiers and the datasets used for both two-class and three-class classification and fine-tuning those three classifiers. 
Section \ref{sec:feat-extract} explains the features extracted from the audio and Section \ref{sec:class-process} describes the classification and hyperparameter optimisation process. 
Section \ref{sec:results} summarises the results and Section \ref{sec:discussion} discusses them. 
Finally, Section \ref{sec:conclusion} concludes this study. 


\section{Data}
\label{sec:data}

We have made use of both public and private data in this study. 
Previously, we have compiled TASK, Sarcos, Brooklyn, and Wallacedene datasets as part of the research projects concerning cough monitoring and cough classification. 
Coswara, ComParE, Google Audio Set \& Freesound and LibriSpeech were compiled from publicly available data. 

Coughs with labels `TB', `COVID-19' and `healthy' are used for the classification task. 
Coughs were excluded from the data used for pre-training altogether as coughs without these three labels may originate from other diseases and we only classified disease in both classification (two-class and \mbox{three-class}) task and fine-tuning the pre-trained DNN classifiers on cough audio. 
All recordings were downsampled to 16 kHz.

	
	

\begin{figure}[h!]
	\centerline{\includegraphics[width=0.5\textwidth]{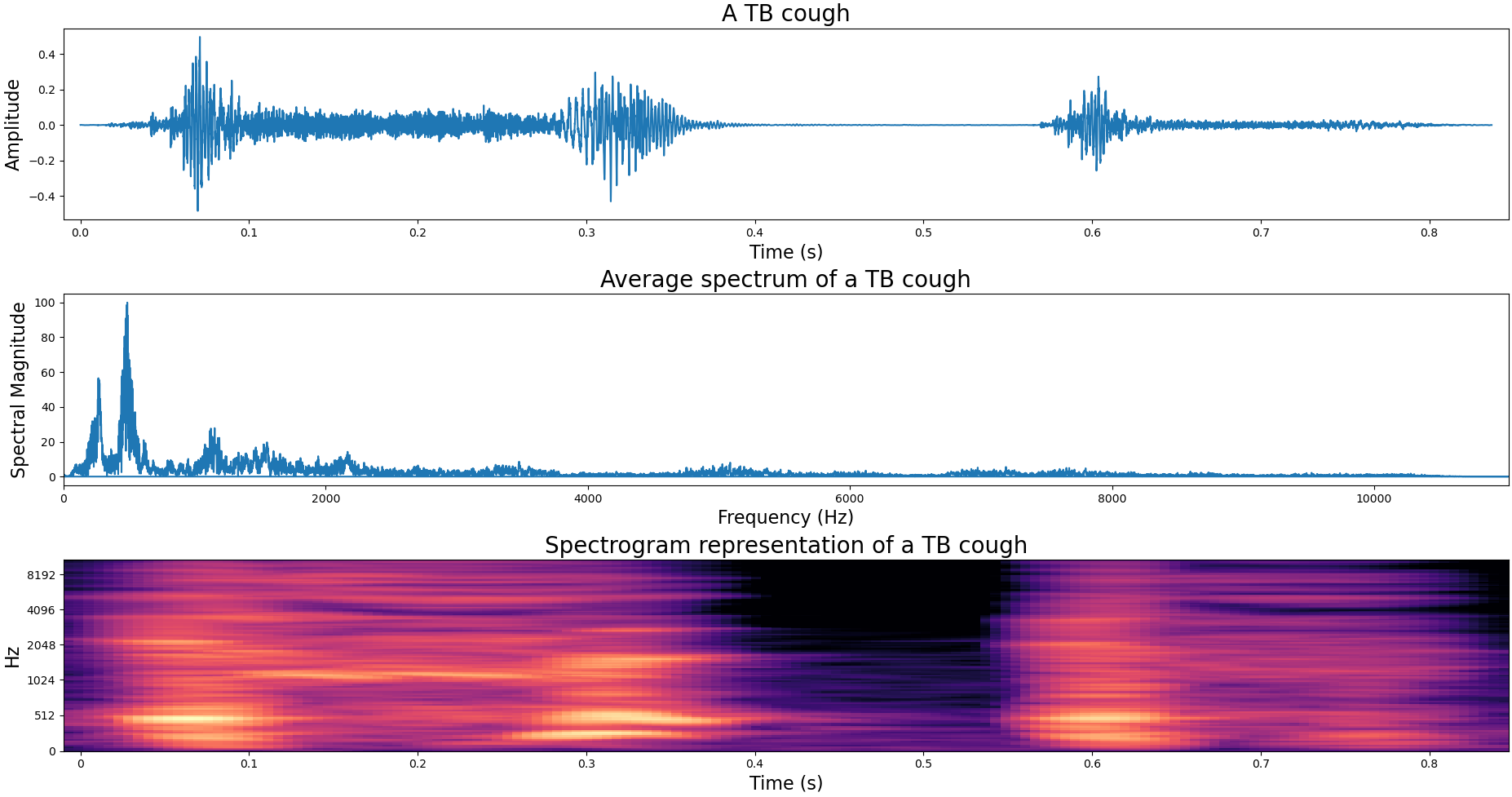}}
	\caption{\textbf{The audio and spectrogram of a TB positive cough.} }
	\label{fig:tb-cough}
\end{figure}

\begin{figure}[h!]
	\centerline{\includegraphics[width=0.5\textwidth]{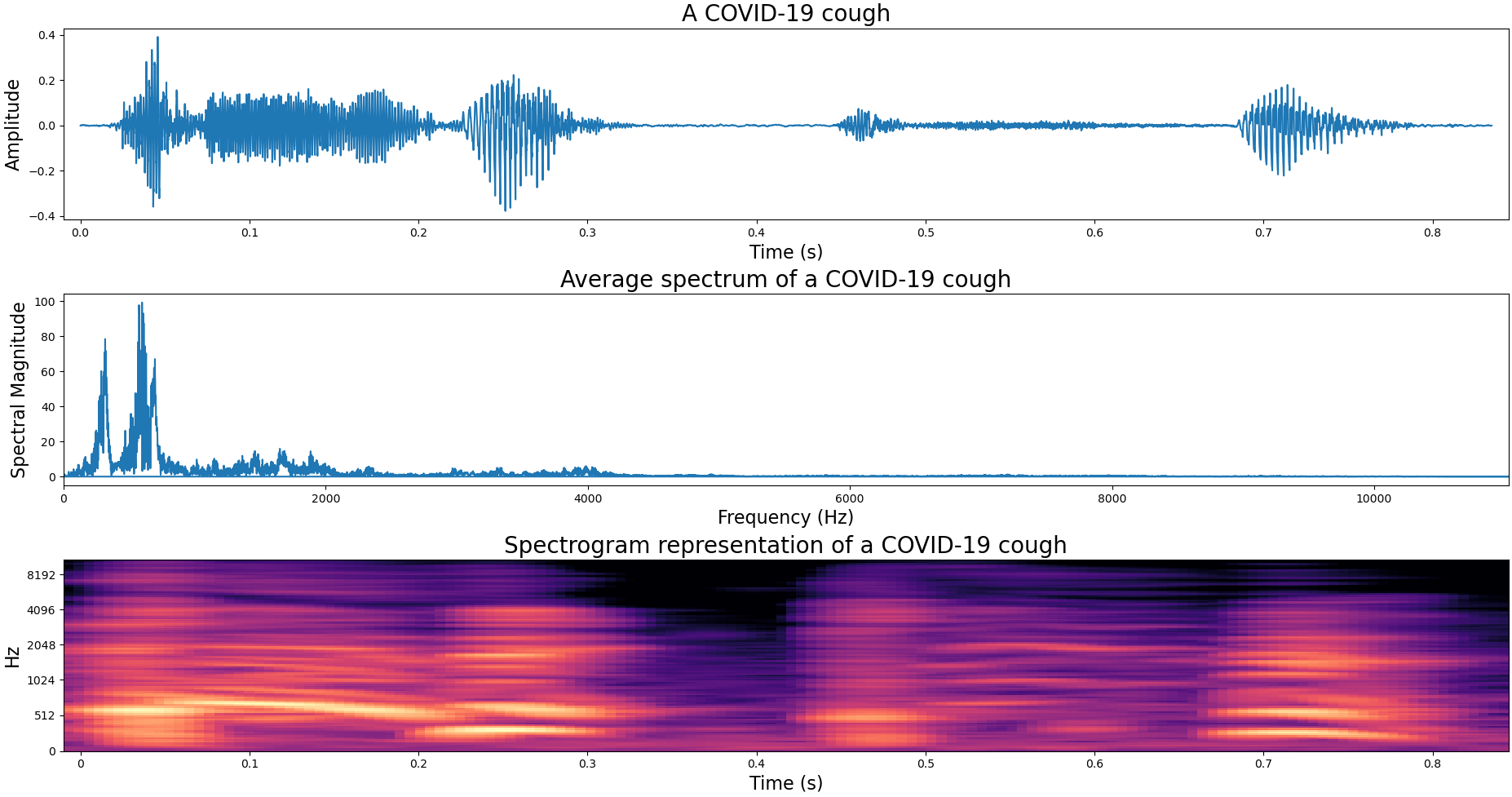}}
	\caption{\textbf{The audio and spectrogram of a COVID-19 positive cough.} }
	\label{fig:covid-cough}
\end{figure}

\begin{figure}[h!]
	\centerline{\includegraphics[width=0.5\textwidth]{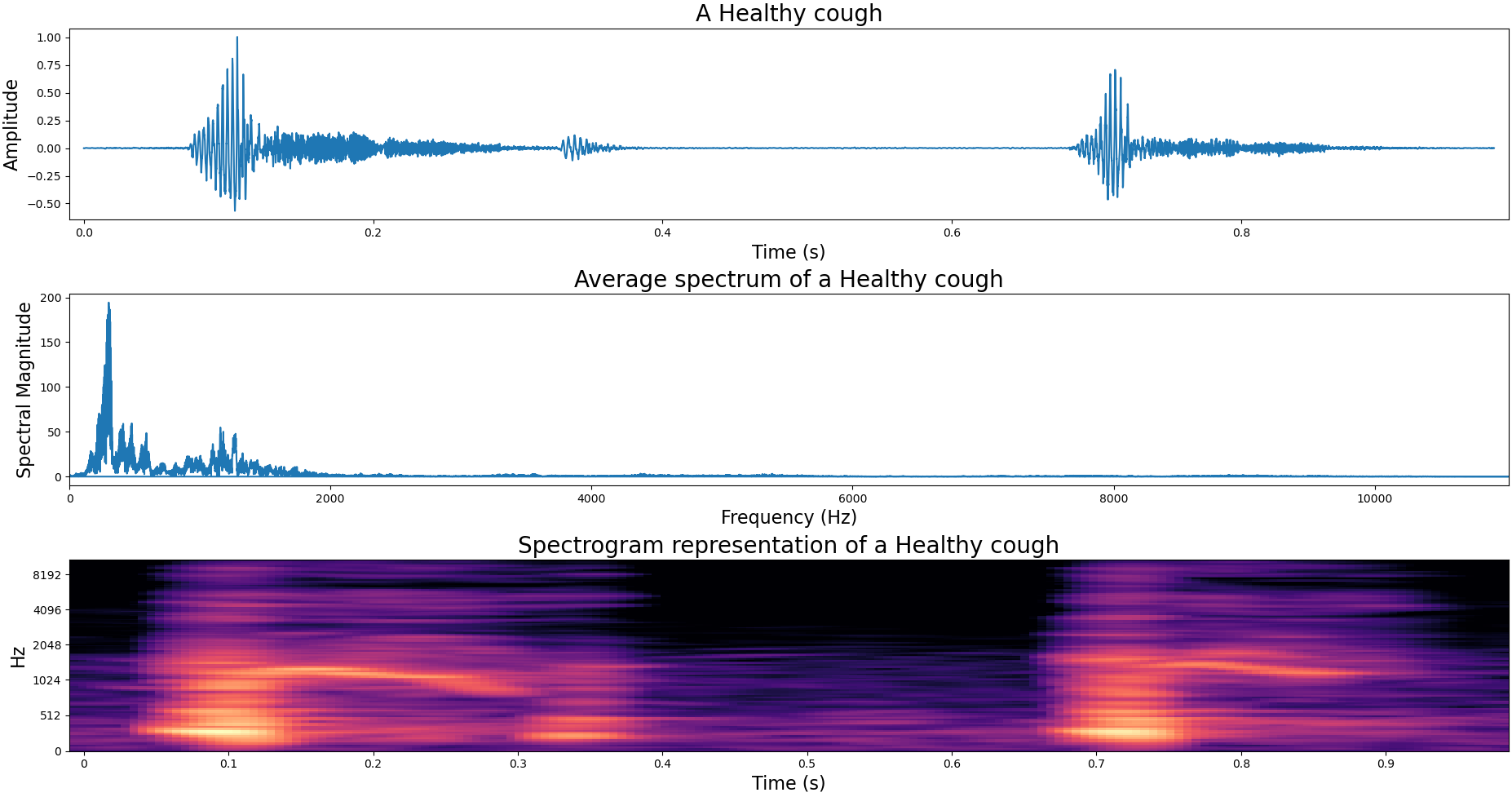}}
	\caption{\textbf{The audio and spectrogram of a healthy cough} which are not much different from those of the TB and COVID-19 coughs, shown in Fig. \ref{fig:tb-cough} and Fig. \ref{fig:covid-cough}. A subjective test couldn't differentiate the sick coughs from the healthy coughs or diagnose the disease. }
	\label{fig:healthy-cough}
\end{figure}

\subsection{Cough audio data for classification}
\label{subsec:TBCoviddata}

The following six datasets of coughs with TB, COVID-19 and healthy labels were available for experimentation and are described in Table \ref{table:class-dataset-summary}.  
A simple energy detector was applied to pre-process the audio recordings of Coswara, Sarcos and ComParE datasets by removing silence within a margin of 50 msec~\cite{pahar2020covid}. 

\subsubsection{TASK dataset}
This dataset contains 6000 continuous cough recordings and 11393 non-cough events such as laughter, doors opening and objects moving.
It was collected at TASK, a TB research centre near Cape Town, South Africa from patients undergoing TB treatment~\cite{pahar2022automatic}. 
Previous research indicates that cough-frequency decreases as patients' health conditions improve \cite{proano2017dynamics}.
Thus, the TASK dataset was compiled to develop cough detection algorithms and monitor patients' long-term health recovery in a multi-bed ward environment using an external microphone attached to a smartphone~\cite{pahar2022accelerometer}. 


\subsubsection{Sarcos dataset}
This dataset was collected in South Africa as part of our own COVID-19 research~\cite{pahar2020covid, pahar2022covid} and contains coughs from 18 COVID-19 positive subjects. 

\subsubsection{Brooklyn dataset}
Cough audio was compiled from 17 TB and 21 healthy subjects to discriminate TB from healthy cough for developing a TB cough audio classifier \cite{botha2018detection}. 
The recordings were taken inside a controlled indoor booth, using an audio field recorder and a R{\O}DE M3 microphone. 

\subsubsection{Wallacedene dataset}
This dataset, containing 402 coughs from 16 TB patients, was collected to extend the previous TB cough audio classification study \cite{botha2018detection} to discriminate TB coughs from other sick coughs in a real-world noisy environment \cite{pahar2021tb}. 
Here, the cough recordings were collected using an audio field recorder and a R{\O}DE M1 microphone and the recording process took place in an outdoor booth, located next to a busy street \cite{pahar2021wake}.

\subsubsection{Coswara dataset}
This publicly available dataset \mbox{(\url{https://coswara.iisc.ac.in})} is compiled to develop machine learning algorithms for the diagnosis of \mbox{COVID-19} in vocal audio~\cite{sharma2020coswara, muguli2021dicova, sharma2022second}. 
Participants from five different continents contributed their vocal audio including coughs using their smartphones. 
In this study, we used the deep coughs from 92 COVID-19 positives and 1079 healthy subjects.

\subsubsection{ComParE dataset}

This dataset was presented in the 2021 Interspeech Computational Paralinguistics ChallengE (ComParE)~\cite{Schuller21-TI2} and it contains 119 COVID-19 positives and 398 healthy subjects.

\subsubsection{Summary of data used for disease classification}
Table~\ref{table:class-dataset-summary} demonstrates that our data contain only 18.54 minutes of COVID-19 cough audio, compared to 1.68 hours of TB coughs and 1.69 hours of healthy coughs, indicating COVID-19 labelled data are under-represented.
As such a data-imbalance can affect the neural networks' performance negatively
~\cite{van2007experimental,krawczyk2016learning}, 
we have applied 
SMOTE~\cite{chawla2002smote}.
This data balancing technique creates new synthetic samples to oversample the minor class
during training. 
SMOTE 
has been successfully applied to address training set class imbalances in cough detection~\cite{pahar2021deep} and cough classification~\cite{pahar2020covid} in the past. 
TASK dataset contains only 14 patients but the length of cough audio per patient was much longer than the other two datasets. 

The audio and spectrograms of a TB, COVID-19 and healthy cough are shown in Fig. \ref{fig:tb-cough}, Fig. \ref{fig:covid-cough} and Fig. \ref{fig:healthy-cough}. 
There are very little obvious visual differences between these three coughs. 
An informal subjective test was conducted where approximately 20 university students were asked to spot the sick and healthy coughs just by listening to these cough audio
and the results showed that human auditory system is unable to spot any disease or differentiate sick coughs from healthy coughs only by listening to the coughs.

\begin{table*}[htbp]
\caption{\textbf{Datasets used in cough classification.} These datasets contain three cough classes: TB, COVID-19 and healthy. } 
\centering 
\begin{center}
	\begin{tabular}{ c | c | c | c | c | c | c }
		\hline
		\hline
		\textbf{Type} & \textbf{Dataset} & \textbf{Sampling rate} & \textbf{No of subjects} & \textbf{Total audio} & \textbf{Average length} & \textbf{Standard deviation}  \\
		\hline
		\hline
		\multirow{4}{*}{TB Cough} & TASK & 44.1 kHz & 14 & 91 mins & 6.5 mins & 1.23 mins \\
		\cline{2-7}
		& Brooklyn & 44.1 kHz & 17 & 4.63 mins & 16.35 sec & 13 sec  \\
		\cline{2-7}
		& Wallacedene & 44.1 kHz & 16 & 4.98 mins & 18.69 sec & 4.95 sec  \\
		\cline{2-7}
		& \textbf{Total (TB Cough)} & \textbf{---} & \textbf{47} & \textbf{1.68 hours} & \textbf{2.14 min} & \textbf{28.37 sec} \\
		
		\hline
		\hline
		\multirow{4}{*}{COVID-19 Cough} & Coswara & 44.1 kHz & 92 & 4.24 mins & 2.77 sec & 1.62 sec  \\
		\cline{2-7}
		& ComParE & 16 kHz & 119 & 13.43 mins & 6.77 sec & 2.11 sec  \\
		\cline{2-7}
		& Sarcos & 44.1 kHz & 18 & 0.87 mins & 2.91 sec & 2.23 sec  \\
		\cline{2-7}
		& \textbf{Total (COVID-19 Cough)} & \textbf{---} & \textbf{229} & \textbf{18.54 mins} & \textbf{4.85 sec} & \textbf{1.92 sec} \\
		
		\hline
		\hline
		\multirow{4}{*}{Healthy Cough} & Coswara & 44.1 kHz & 1079 & 0.98 hours & 3.26 sec & 1.66 sec  \\
		\cline{2-7}
		& ComParE & 16 kHz & 398 & 40.89 mins & 6.16 sec & 2.26 sec  \\
		\cline{2-7}
		& Brooklyn & 44.1 kHz & 21 & 1.66 mins & 4.7 sec & 3.9 sec  \\
		\cline{2-7}
		& \textbf{Total (Healthy Cough)} & \textbf{---} & \textbf{1498} & \textbf{1.69 hours} & \textbf{4.05 sec} & \textbf{1.85 sec} \\
		
		\hline
		\hline
		
	\end{tabular}
\end{center}
\label{table:class-dataset-summary}
\end{table*}

\subsection{Datasets without cough labels for pre-training}
\label{subsec:pretraindata}

Our classifier training is limited as cough audio data is not available abundantly.
Hence, we use three other types of audio data for pre-training and they include sneeze, speech and noise from Google Audio Set \& Freesound, LibriSpeech and TASK datasets, as described in Table \ref{table:pre-train-dataset-summary}.

\subsubsection{Google Audio Set \& Freesound}
The Google Audio Set has manually-labelled excerpts from 1.8 million Youtube videos belonging to 632 audio event categories~\cite{gemmeke2017audio}. 
The Freesound audio database contains tagged audio with various noise levels, uploaded by subjects from many parts of the world under widely varying recording conditions~\cite{font2013freesound}. 
From these audio, We have selected the 
recordings that include 1013 sneezes, 2326 speech excerpts and 1027 other non-vocal sounds such as restaurant chatter, running water and engine noise. 
This manually annotated dataset was successfully used in developing 
cough detection algorithms~\cite{miranda2019comparative}. 

\subsubsection{LibriSpeech}

The LibriSpeech corpus~\cite{panayotov2015librispeech} is freely available and contains very little noise.
We have carefully selected utterances from 28 female and 28 male speakers.


\subsubsection{Summary of data used for pre-training}

In total, the data described in Table \ref{table:pre-train-dataset-summary} includes 1013 sneezing sounds (13.34 minutes of audio), 2.91 hours of speech,  and 2.98 hours of noise. 
As sneezing is under-represented,
we have again applied 
SMOTE to create additional synthetic samples.
In total, a dataset containing  7.84 hours of audio recordings with three class labels (sneeze, speech, noise) was used to pre-train three DNN classifiers.

\begin{table*}[htbp]
\caption{\textbf{Datasets used in pre-training.} DNN classifiers are pre-trained on 7.84 hours of audio data with three class labels: sneeze, speech and noise. These data do not contain any cough. }  
\centering 
\begin{center}
	\begin{tabular}{ c | c | c | c | c | c | c }
		\hline
		\hline
		\textbf{Type} & \textbf{Dataset} & \textbf{Sampling rate} & \textbf{No of events} & \textbf{Total audio} & \textbf{Average length} & \textbf{Standard deviation}  \\
		
		\hline \hline
		\multirow{3}{*}{Sneeze} & Google Audio Set \& Freesound & 16 kHz & 1013 & 13.34 mins & 0.79 sec & 0.21 sec  \\
		\cline{2-7}
		& Google Audio Set \& Freesound + SMOTE & 16 kHz & 9750 & 2.14 hours & 0.79 sec & 0.23 sec \\
		\cline{2-7}
		& \textbf{Total (Sneeze)} & \textbf{---} & \textbf{10763} & \textbf{2.14 hours} & \textbf{0.79 sec} & \textbf{0.23 sec} \\
		
		\hline \hline
		\multirow{3}{*}{Speech} & Google Audio Set \& Freesound & 16 kHz & 2326 & 22.48 mins & 0.58 sec & 0.14 sec  \\
		\cline{2-7}
		& LibriSpeech & 16 kHz & 56 & 2.54 hours & 2.72 mins & 0.91 mins  \\
		\cline{2-7}
		& \textbf{Total (Speech)} & \textbf{---} & \textbf{2382} & \textbf{2.91 hours} & \textbf{4.39 sec} & \textbf{0.42 sec} \\
		
		\hline \hline
		\multirow{3}{*}{Noise}& TASK dataset & 44.1 kHz & 12714 & 2.79 hours & 0.79 sec & 0.23 sec \\
		\cline{2-7}
		& Google Audio Set \& Freesound & 16 kHz & 1027 & 11.13 mins & 0.65 sec & 0.26 sec \\
		\cline{2-7}
		& \textbf{Total (Noise)} & \textbf{---} & \textbf{13741} & \textbf{2.79 hours} & \textbf{0.79 sec} & \textbf{0.23 sec} \\
		\hline
		\hline
		
	\end{tabular}
\end{center}
\label{table:pre-train-dataset-summary}
\end{table*}

\section{Feature Extraction}
\label{sec:feat-extract}

Mel-frequency cepstral coefficients (MFCCs) along with their velocity and acceleration coefficients, zero-crossing rate
and kurtosis
~\cite{bachu2010voiced} 
were extracted from the audio recordings and these features were used for both classification and \mbox{pre-training} task. 
The feature combination containing MFCCs rather than linearly-spaced log filerbanks \cite{botha2018detection} showed better performance in our previous TB \cite{pahar2021tb} and COVID-19 \cite{pahar2020covid, pahar2021deep} classification tasks.
MFCCs are the features of choice in detecting and classifying voice audio such as speech \cite{paharrecreating, pahar_coding_2020, pahar2012reconstructing} and coughs \cite{miranda2019comparative}.

Overlapping frames were used to extract features, where the frame overlap ensures that a certain exact number of frames always represents the entire audio event. 
This way an \mbox{image-like} fixed input dimension feature matrix can be computed where the general overall temporal structure of the sound are also maintained. 
Such fixed two-dimensional features have been successfully used to train DNN classifiers in our previous experiments \cite{pahar2020covid, pahar2021deep}. 

The dimension of the input feature matrix has been \mbox{($3\mathbb{M} + 2, \mathbb{S}$)} for $\mathbb{M}$ MFCCs.
The frame length ($\mathbb{F}$), exact number of frames ($\mathbb{S}$) and number of lower order MFCCs ($\mathbb{M}$) are used as the feature extraction hyperparameters, mentioned in Table~\ref{table:feat-hyper-parameter}. 
The table shows that $\mathbb{M}$ lies between 13 and 65, which varies the spectral information in each audio event and 
each audio event is divided into between 70 and 200 frames, as different phases of coughs carry different information.
Each frame consists of between 512 and 4096 samples, i.e. 32 msec and 256 msec of audio, as the sampling rate is 16 kHz in our experiments.

\begin{table}[htbp]
\footnotesize
\setlength{\tabcolsep}{3pt} 
\caption{\textbf{Feature extraction hyperparameters.} By varying the MFCCs between 13 \& 65, frame lengths between 512 \& 4096 and no of frames between 70 \& 200, the spectral resolutions of the audio have been varied over a large range.  } 
\centering 
\begin{center}
	\begin{tabular}{ c | c | c }
		\hline
		\hline
		\textbf{Hyperparameters} & \textbf{Description}     & \textbf{Range} \\
		\hline
		\hline
		\multirow{2}{*}{MFCCs ($\mathbb{M}$)}  & \multirow{2}{*}{lower order MFCCs to keep}   & $13 \times k$, where \\
		&  & $k=1, 2, 3, 4, 5$  \\
		\hline
		\multirow{2}{*}{Frame length ($\mathbb{F}$)}  & \multirow{2}{*}{into which audio is segmented} & $2^{k}$, where \\
		&  & $k=9, 10, 11, 12$ \\
		\hline
		\multirow{2}{*}{Segments ($\mathbb{S}$)}    & \multirow{2}{*}{no. of frames extracted from audio} & $10 \times k$, where \\
		&  & $k=7, 10, 12, 15$ \\
		\hline
		\hline
	\end{tabular}
\end{center}
\label{table:feat-hyper-parameter}
\end{table}

\section{Classification Process}
\label{sec:class-process}

\subsection{Classifier Architectures}

We have used only three DNN classifiers: CNN\cite{krizhevsky2017imagenet}, LSTM \cite{hochreiter1997long} and Resnet50 \cite{he2016deep} in this study. 
We have refrained from experimenting with any shallow classifier as using deep architectures along with SMOTE data balancing technique yielded better results in our previous experiments \cite{pahar2020covid, pahar2021deep}. 
For our initial set of experiments, we have used these three DNN classifiers for two-class (TB vs COVID-19) and \mbox{three-class} (TB vs COVID-19 and healthy) classification and the classifier hyperparameters are mentioned in Table \ref{table:class-hyper-parameter}. 
The classifier training process is stopped when the performance wasn't improved after 10 epochs. 
Finally, for the improved performance, we have applied the transfer learning.

\begin{table}[htbp]
\footnotesize
\caption{\textbf{Classifier hyperparameters}, optimised using leave-$p$-out nested cross-validation scheme. }
\centering 
\begin{center}
	\begin{tabular}{ l | c | l  }
		\hline
		\hline
		\textbf{Hyperparameters} & \textbf{Classifier} & \textbf{Range} \\
		\hline
		\hline
		No. of conv filters ($\alpha_1$) & CNN & $3 \times 2^k$ where $k=3, 4, 5$ \\
		\hline
		Kernel size ($\alpha_2$) & CNN & 2 and 3 \\
		\hline
		Dropout rate ($\alpha_3$) & CNN, LSTM  & 0.1 to 0.5 in steps of 0.2 \\
		\hline
		Dense layer units ($\alpha_4$) & CNN, LSTM  & $2^k$ where $k=4, 5$ \\
		\hline
		LSTM units ($\alpha_5$) & LSTM & $2^k$ where $k=6, 7, 8$ \\
		\hline
		Learning rate ($\alpha_6$) & LSTM & $10^k$ where, $k=-2,-3,-4$ \\
		\hline
		Batch Size ($\alpha_7$) & CNN, LSTM  & $2^k$ where $k=6, 7, 8$\\
		\hline
		\hline
	\end{tabular}
\end{center}
\label{table:class-hyper-parameter}
\end{table}

\subsection{Transfer Learning Architectures}

\begin{figure*}[h!]
\centerline{\includegraphics[width=0.99\textwidth]{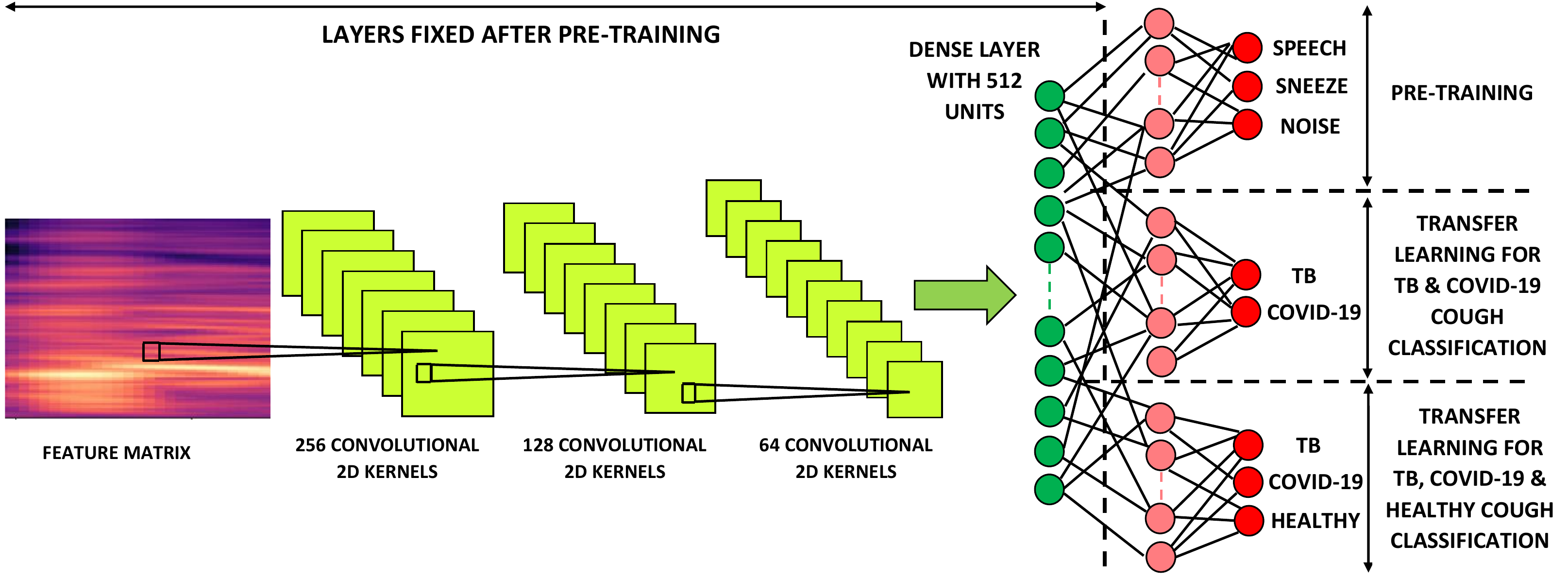}}
\caption{\textbf{Transfer learning architecture for the CNN:} Cross-validation on the pre-training data produced optimal results when three convolutional layers (256, 128 \& 64) with (2 $\times$ 2) kernels were used, followed by (2, 2) max-pooling. The outputs of these three convolutional layers were flattened and passed through two fully connected layers (with a dropout rate of 0.3), each consisting 512, 128 relu units. The final fully connected layer consists of 3 softmax units. To apply transfer learning, the final two layers were taken away and was replaced by two fully connected layers with 16 and 2 units for two-class (TB and COVID-19) cough classification and with 16 and 3 units for three-class (TB, COVID-19 and healthy) cough classification. 
}
\label{fig:class-arch}
\end{figure*}

The application of transfer learning has improved the classification performance in our previous studies \cite{pahar2020covid, pahar2022covid}. 
Hence, we have also applied transfer learning in this study to improve the classification performance, where the DNN classifiers are pre-trained on the dataset, explained in Section \ref{subsec:pretraindata}, and then fine-tuned on the classification datasets, explained in Section \ref{subsec:TBCoviddata}. 
The feature extraction hyperparameters are adopted from our previous studies \cite{pahar2020covid, pahar2022covid} and the hyperparameters of the CNN and LSTM were determined during the cross-validation process. 
These hyperparameters are mentioned in Table \ref{table:pre-train-hyper-parameter}. 
A standard Resnet50, as explained in Table 1 of \cite{he2016deep}, with \mbox{512-unit} dense layer has been used for the transfer learning.
The transfer learning process for a CNN is explained in Fig. \ref{fig:class-arch}.

\begin{table}[htbp]
\footnotesize
\caption{\textbf{Feature extraction and classifier hyperparameters of the pre-trained networks:} We used the same feature extraction hyperparameters used in our previous work~\cite{pahar2020covid, pahar2022covid}, while classifier hyperparameters were optimised on the pre-training data (Table \ref{table:pre-train-dataset-summary}) using the nested cross-validation. }
\centering 
\begin{center}
	\begin{tabular}{ c | c | c  }
		
		\hline
		\hline
		\multicolumn{3}{c}{\textbf{\uppercase{Feature Extraction hyperparameters}}}\\
		\hline
		\hline
		
		\multicolumn{2}{c|}{\textbf{Hyperparameters}} & \textbf{Values} \\
		
		\hline
		$\mathbb{M}$   & MFCCs & $39$ \\
		
		\hline
		$\mathbb{F}$  & frame length  & $2^{10} = 1024$ \\
		
		\hline
		$\mathbb{S}$  & no. of frames & $150$ \\

		\hline
		\hline
		\multicolumn{3}{c}{\textbf{\uppercase{Classifier hyperparameters}}}\\
		\hline
		\hline
		\textbf{Hyperparameters} & \textbf{Classifier} & \textbf{Values} \\
		\hline
		No. of conv filters ($\alpha_1$) & CNN & $256$ \& $128$ \& $64$ \\
		\hline
		Kernel size ($\alpha_2$) & CNN & $2$ \\
		\hline
		Dropout rate (($\alpha_3$)) & CNN, LSTM  & $0.3$ \\
		\hline
		Dense layer units ($\alpha_4$) & \multirow{2}{*}{CNN, LSTM, Resnet50} & \multirow{2}{*}{$512$ \& $128$ \& $3$} \\[-0.2em]
		(for pre-training) &  &  \\
		\hline
		Dense layer units ($\alpha_4$) & \multirow{2}{*}{CNN, LSTM, Resnet50} & \multirow{2}{*}{$16$ \& $2$ or $3$} \\[-0.2em]
		(for fine-tuning) &  &  \\
		\hline
		LSTM units ($\alpha_5$) & LSTM & $512$ \& $256$ \& $128$ \\
		\hline
		Learning rate ($\alpha_6$) & LSTM & $10^{-3} = 0.001$ \\
		\hline
		Batch Size ($\alpha_7$) & CNN, LSTM, Resnet50  & $2^7 = 128$ \\
		\hline
		\hline
	\end{tabular}
\end{center}
\label{table:pre-train-hyper-parameter}
\end{table}

\subsection{Hyperparameter optimisation}

The feature extraction process and classifiers have a number of hyperparameters, listed in Table \ref{table:feat-hyper-parameter} and \ref{table:class-hyper-parameter}. 
They were optimised by using a leave-$p$-out cross-validation scheme \cite{liu2019leave}. 
The train and test split ratio was 4:1, due to its effectiveness in medical applications \cite{racz2021effect}. 
This 5-fold cross-validation process ensured the best use of our dataset by using all subjects in both training and testing the classifiers and implementing a strict no patient-overlap between cross-validation folds.

\subsection{Classifier evaluation}

The F1-score has been the optimisation criteria in the \mbox{cross-validation} folds and the performance-indicator of the classifiers \cite{fourure2021anomaly}.
We note the mean per-frame probability that a cough is from a COVID-19 positive subject is $\hat{C}$ in Equation \ref{eq:P-hat}. 
\begin{equation}
\hat{C} = \frac{\sum\limits_{i=1}^{\mathbb{S}} P(Y = 1|X_i, \theta)}{\mathbb{S}} 
\label{eq:P-hat}
\end{equation} 
where $P(Y = 1|X_i, \theta)$ is the output of the classifier for feature vector $X_i$ and parameters $\theta$ for the $i^{th}$ frame. 







The average F1-score along with its 
standard deviation ($\sigma_{F1}$) over the outer folds during cross-validation are shown in Tables \ref{table:TB-COVID-results}, \ref{table:TB-COVID-Healthy-results}.  
Hyperparameters producing the highest \mbox{F1-score} over the inner loops of the cross-validation scheme have been noted as the `best classifier hyperparameters' in Tables \ref{table:TB-COVID-results}, \ref{table:TB-COVID-Healthy-results}.

\section{Results}
\label{sec:results}

\subsection{TB and COVID-19 cough classification}

\begin{figure}[h]
\centerline{\includegraphics[width=0.51\textwidth]{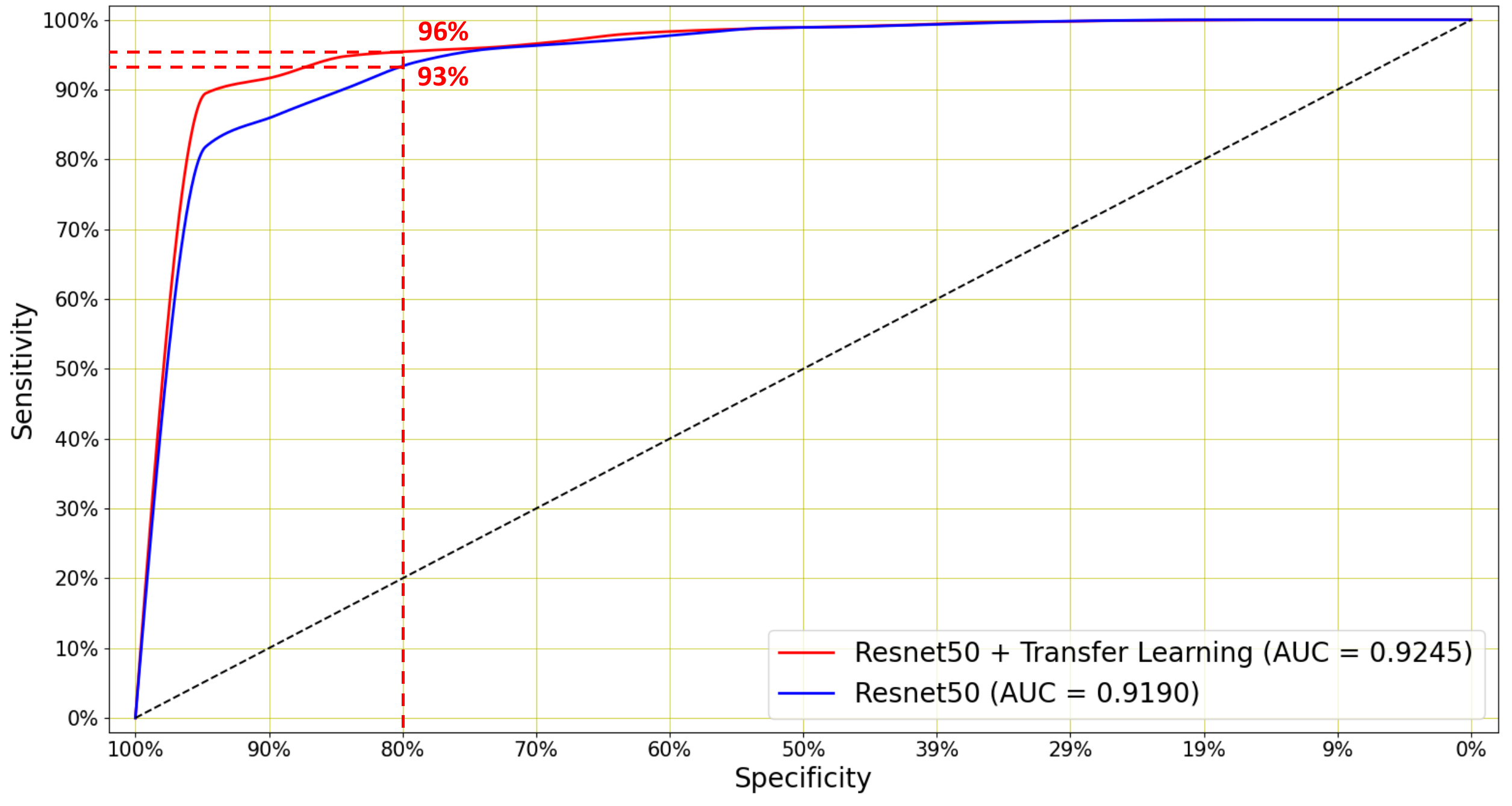}}
\caption{\textbf{The ROC curves for discriminating TB coughs from COVID-19 coughs:} An AUC of 0.9190 is achieved from the Resnet50 and the highest AUC of 0.9245 is achieved after applying transfer learning to this Resnet50 architecture. Both systems achieve 96\% and 93\% sensitivity respectively at 80\% specificity, thus exceed the community-based TB Triage test requirement of 90\% sensitivity at 70\% specificity set by WHO. }
\label{fig:TB-COVID-AUC}
\end{figure}

\begin{table*}[htbp]
\centering
\caption{\textbf{Classifying TB and COVID-19 coughs:} Resnet50 has performed the best in discriminating TB coughs from COVID-19 coughs. The initial experiment achieved the F1-score of 0.9042 and the AUC of 0.9190, along with the $\sigma_{F1}$ of 0.83. After applying the transfer learning, F1-score and AUC increase to 0.9259 and 0.9245 and $\sigma_{F1}$ decreases to 0.03. }
\begin{tabular}{ c | c | c | c | c | c }
	\hline
	\hline
	{\multirow{2}{*}{\textbf{Classifier}}} & \textbf{Best Feature} &
	\textbf{Best Classifier Hyperparameters} & \multicolumn{3}{c}{\textbf{Performance}} \\
	
	\cline{4-6}
	
	& \textbf{Hyperparameters} & \textbf{(Optimised inside nested cross-validation)} & \textbf{F1-score} & \textbf{$\sigma_{F1}$} & \textbf{AUC} \\
	
	\hline
	\hline
	Resnet50 & $\mathbb{M}=26, \mathbb{F}=1024, \mathbb{S}=150$ & Default Resnet50 (Table 1 in \cite{he2016deep}) & 0.9042 & 83$\times 10^{-2}$ & 0.9190 \\
	\hline
	CNN & $\mathbb{M}=26, \mathbb{F}=2048, \mathbb{S}=100$ & $\alpha_1$=256, $\alpha_2$=2, $\alpha_3$=0.3, $\alpha_4$=32, $\alpha_7$=256 & 0.8887 & 61$\times 10^{-2}$ & 0.8895 \\
	\hline
	LSTM & $\mathbb{M}=39, \mathbb{F}=2048, \mathbb{S}=120$ & $\alpha_3$=0.1, $\alpha_4$=32, $\alpha_5$=128, $\alpha_6$=0.001, $\alpha_7$=256 & 0.8802 & 49$\times 10^{-2}$ & 0.8884 \\
	
	\hline
	\hline
	\textit{Resnet50 + Transfer Learning} & \textit{Table \ref{table:pre-train-hyper-parameter}} & \textit{Default Resnet50 (Table 1 in \cite{he2016deep})} & \textit{0.9259} & \textit{3$\times 10^{-2}$} & \textit{0.9245} \\
	\hline
	LSTM + Transfer Learning & \dittoclosing & Table \ref{table:pre-train-hyper-parameter} & 0.9134 & 4$\times 10^{-2}$ & 0.9124 \\
	\hline
	CNN + Transfer Learning & \dittoclosing & \dittoclosing & 0.9127 & 4$\times 10^{-2}$ & 0.9211 \\

	\hline
	\hline
	
\end{tabular}
\label{table:TB-COVID-results}
\end{table*}

\begin{table*}[htbp]
\centering
\caption{\textbf{Classifying TB and COVID-19 and healthy coughs:} Resnet50 has again been the classifier of the choice by producing the highest F1-score of 0.8578 with a $\sigma_{F1}$ of 0.67. This performance has been improved to an F1-score of 0.8631 with a lower $\sigma_{F1}$ of 0.11 after applying the transfer learning. }
\begin{tabular}{ c | c | c | c | c | c }
	\hline
	\hline
	{\multirow{2}{*}{\textbf{Classifier}}} & \textbf{Best Feature} &
	\textbf{Best Classifier Hyperparameters} & \multicolumn{3}{c}{\textbf{Performance}} \\
	
	\cline{4-6}
	
	& \textbf{Hyperparameters} & \textbf{(Optimised inside nested cross-validation)} & \textbf{F1-score} & \textbf{$\sigma_{F1}$} & \textbf{Accuracy} \\
	
	\hline
	\hline
	Resnet50 & $\mathbb{M}=39, \mathbb{F}=1024, \mathbb{S}=120$ & Default Resnet50 (Table 1 in \cite{he2016deep}) & 0.8578 & 67$\times 10^{-2}$ & 0.8662 \\
	\hline
	CNN & $\mathbb{M}=26, \mathbb{F}=1024, \mathbb{S}=150$ & $\alpha_1$=256, $\alpha_2$=2, $\alpha_3$=0.3, $\alpha_4$=16, $\alpha_7$=128 & 0.8220 & 41$\times 10^{-2}$ & 0.8311 \\
	\hline
	LSTM & $\mathbb{M}=26, \mathbb{F}=2048, \mathbb{S}=120$ & $\alpha_3$=0.1, $\alpha_4$=32, $\alpha_5$=128, $\alpha_6$=0.001, $\alpha_7$=256 & 0.8125 & 49$\times 10^{-2}$ & 0.8181 \\
	
	\hline
	\hline
	\textit{Resnet50 + Transfer Learning} & \textit{Table \ref{table:pre-train-hyper-parameter}} & \textit{Default Resnet50 (Table 1 in \cite{he2016deep})} & \textit{0.8631} & \textit{11$\times 10^{-2}$} & \textit{0.8689} \\
	\hline
	CNN + Transfer Learning & \dittoclosing & Table \ref{table:pre-train-hyper-parameter} & 0.8455 & 7$\times 10^{-2}$ & 0.8564 \\
	\hline
	LSTM + Transfer Learning & \dittoclosing & \dittoclosing & 0.8427 & 9$\times 10^{-2}$ & 0.8490 \\
	
	\hline
	\hline
	
\end{tabular}
\label{table:TB-COVID-Healthy-results}
\end{table*}

For the initial classification task in Table \ref{table:TB-COVID-results}, the Resnet50 architecture has performed the best by producing the highest mean F1-score of 0.9042 and mean AUC of 0.9190 with the $\sigma_{F1}$ of 0.83. 
Although the CNN and LSTM have produced a lower F1-score and AUC, the $\sigma_{F1}$ has also been lower, 0.61 and 0.49 respectively, suggesting better generalisation and robustness over the folds for less deep architectures. 
This also indicates that the very deep architectures such as a Resnet50, although able to perform better, are prone to over-fitting. 
The best feature hyperparameters have been 26 MFCCs, 1024 sample-long frames and 150 frames per event such as a cough. 

To prevent over-fitting, we have applied transfer learning and noticed a slight improvement in the DNN classifiers' performance. 
The F1-score and the AUC have increased to 0.9259 and 0.9245 and the $\sigma_{F1}$ has decreased to 0.03 from the pre-trained Resnet50 classifier. 
A similar trend has also been noticed for CNN and LSTM classifiers, where their performance (F1-score and AUC) has also increased along with a lower $\sigma_{F1}$. 

Although the CNN outperformed the LSTM initially, LSTM has outperformed the CNN after applying transfer learning. 
The mean ROC curves for the initial and pre-trained Resnet50 are shown in Fig. \ref{fig:TB-COVID-AUC}. 
These two systems achieve 96\% and 93\% sensitivity respectively at 80\% specificity.
Thus they exceed the community-based TB Triage test requirement of 90\% sensitivity at 70\% specificity set by WHO \cite{world2014high}.

\subsection{TB and COVID-19 and healthy cough classification}

We observe a similar pattern in three-class classification as well.  
Table \ref{table:TB-COVID-Healthy-results} shows that the highest F1-score of 0.8578 has been achieved from the Resnet50 classifier with a $\sigma_{F1}$ of 0.67 from the best feature hyperparameters of 39 MFCCs, 1024 sample-long frames and 120 frames per cough. 
At the same time, CNN and LSTM produce the F1-scores of 0.8220 and 0.8125 with $\sigma_{F1}$ of 0.41 and 0.49 respectively. 
Both these F1-scores and $\sigma_{F1}$ scores are lower than those produced by the Resnet50. 
As this is a three-class classification, we have replaced AUC with accuracy in Table \ref{table:TB-COVID-Healthy-results}. 
Again, the signs of overfitting are clear in these performances and we apply transfer learning next. 

Application of the transfer learning has improved the classification performance by a small margin. 
The F1-score from the Resnet50 rose to 0.8631 and the $\sigma_{F1}$ decreased to 0.11. 
The performances from CNN and LSTM have also improved, as the F1-scores of 0.8455 and 0.8427 have been achieved from these two DNN classifiers, respectively. 
Their $\sigma_{F1}$ scores are also much lower: 0.07 and 0.09 respectively. 
Although, pre-trained CNN and LSTM models have produced lower \mbox{F1-scores}, their $\sigma_{F1}$ is also lower, unlike in the previous \mbox{two-class} classification. 
This shows that the application of transfer learning helps the classifiers to be more robust in classification tasks.

\section{Discussion}
\label{sec:discussion}

Although many previous studies have shown that both TB and COVID-19 can be discriminated from healthy coughs, here we show that there are unique disease signatures present in cough audio which is responsible for the machine learning classifiers to discriminate TB coughs from COVID-19 coughs. 
We have experimentally found that when the cough data are limited, classifier performance can be poor and they are also prone to overfitting. 
Very deep architectures generally produce higher mean F1-scores, however with the expense of higher variances along the cross-validation folds. 
Our study shows that the application of transfer learning using vocal data which do not even include cough can be used to improve classifiers' performance in disease classification. 

TB and COVID-19 are the two most deadly respiratory diseases transmitted via droplets that are coughed out. 
Thus, a \mbox{contact-less} diagnosis using a smartphone would be the most desirable solution in these conditions, as opposed to other common coughs from allergic asthma, chronic obstructive pulmonary disease (COPD), bronchitis, common colds, etc, that are not contagious. 
The deep learning classifiers presented in this study can be implemented in a smartphone, thus enabling the diagnosis process fully non-contact and without needing any expensive laboratory testing equipment, thus protecting the environment and the health care professionals from possible exposure to health risks.

\section{Conclusion and Future Work}
\label{sec:conclusion}

Here in this study, a deep learning based cough classifier which can discriminate between TB coughs and \mbox{COVID-19} coughs and healthy coughs has been presented, where a subjective test confirmed that respiratory disease can't be confirmed just by listening to the cough audio. 
The cough audio recordings contain various types and levels of background noise as they were collected inside a TB research centre, recording booth and by using smartphones from subjects around the globe. 
This cough data include 47 TB subjects, 229 COVID-19 subjects and 1498 healthy subjects contributing 1.68 hours, 18.54 minutes and 1.69 hours of audio respectively. 
Application of transfer learning has yielded better performance in our previous studies, thus a separate data containing 2.14 hours of sneeze, 2.91 hours of speech and 2.79 hours of noise such as door slamming, engine running, etc, have been used to pre-train three deep neural networks: CNN, LSTM and Resnet50. 
The class-imbalance in our dataset was addressed by using SMOTE data balancing technique during the training process and using performance metrics such as F1-score and AUC. 
The classifiers were evaluated by using a 5-fold nested cross-validation scheme. 
The experimental results show that the highest F1-score of 0.9259 has been achieved from a \mbox{pre-trained} Resnet50 for the two-class (TB vs COVID-19) cough classification task and the highest F1-score of 0.8631 has been achieved from a pre-trained Resnet50 for three-class (TB vs COVID-19 vs healthy) cough classification task. 
The pre-trained Resnet50 architecture also produces the highest AUC of 0.9245 with 96\% sensitivity at 80\% specificity, which exceeds the TB triage test requirement of 90\% at 70\% specificity. 
The results also show that the application of transfer learning has improved the performance and generalises better over the cross-validation folds, making the classifiers more robust. 
The best feature hyperparameters also contain higher order of MFCCs, suggesting auditory patterns responsible for disease classification are not perceivable by the human auditory system. 
This type of cough audio classification is non-contact, cost-effective and can easily be deployed to a smartphone, thus it can be a useful tool for an automatic non-invasive TB and COVID-19 screening, especially in a developing country setting, where these two most deadly contagious diseases claim thousands of lives each year. 

As for the future work, we are investigating the length of the coughs required for effective overall classification scores. 
We are also compiling a bigger dataset containing both TB and COVID-19 patients to improve the existing cough classification models and deploying the TensorFlow-based models on an Android and iOS platform.

\section*{Acknowledgment}

This research was supported by the South African Medical Research Council (SAMRC) through its Division of Research Capacity Development under the SAMRC Intramural Postdoctoral programme and the Research Capacity Development Initiative as well as the COVID-19 IMU EMC allocation from funding received from the South African National Treasury.
We also acknowledge funding from the EDCTP2 programme supported by the European Union (grant SF1401, OPTIMAL DIAGNOSIS; grant RIA2020I-3305, CAGE-TB) and the National Institute of Allergy and Infection Diseases of the National Institutes of Health (U01AI152087).

We would like to thank the South African Centre for High Performance Computing (CHPC) for providing computational resources on their Lengau cluster for this research and gratefully acknowledge the support of Telkom South Africa. 
We also thank the Clinical Mycobacteriology \& Epidemiology (CLIME) clinic team for assisting in data collection, especially Sister Jane Fortuin and Ms. Zintle Ntwana. 
We also especially thank Igor Miranda, Corwynne Leng, Renier Botha, Jordan Govendar and Rafeeq du Toit for their support in data collection and annotation. 

The content and findings reported are the sole deduction, view and responsibility of the researcher and do not reflect the official position and sentiments of the SAMRC, EDCTP2, European Union or the funders.


\bibliographystyle{IEEEtran}
\bibliography{reference}

\end{document}